\documentclass[letterpaper,10pt,conference]{ieeeconf}

\IEEEoverridecommandlockouts
\usepackage{amsmath}
\usepackage{amssymb}
\usepackage{graphicx}
\usepackage{booktabs}
\usepackage{multirow}
\usepackage{cite}
\usepackage{xcolor}
\usepackage{colortbl}
\usepackage{xspace}
\usepackage{balance}
\usepackage{array}
\usepackage{url}
\usepackage{siunitx}
\usepackage{bm}
\usepackage{xcolor}
\definecolor{stdgray}{gray}{0.55}
\usepackage[table]{xcolor}
\usepackage{makecell}

\newcommand{\method}{\textsc{JAMB}\xspace}

\newcommand{\result}[2]{%
  #1\,{\color{stdgray}\scriptsize $\pm #2$}%
}
\newcommand{\bestresult}[2]{%
  \textbf{#1}\,{\color{stdgray}\scriptsize $\pm #2$}%
}

\title{\LARGE \bf
JAMB: Joint Action--Motion Diffusion for Bimanual Manipulation
}

\author{
Chuyang Xiao$^{1,*}$, Peilin Meng$^{2,*}$, David Held$^{1,\dagger}$%
\thanks{$^{1}$Robotics Institute, Carnegie Mellon University,
Pittsburgh, PA 15213, USA.}%
\thanks{$^{2}$University of Michigan,
Ann Arbor, MI 48109, USA.}%
\thanks{$^{*}$Equal contribution. $^{\dagger}$Corresponding author.}
}

\begin{document}
\maketitle
\thispagestyle{empty}
\pagestyle{empty}

\begin{abstract}
Coordinated bimanual manipulation is challenging because the motion of either arm can alter the shared 3D scene and thereby affect the other arm. Yet most diffusion policies generate actions without explicitly modeling these future geometric consequences, while predictive variants typically use future state only as auxiliary supervision or fixed conditioning. We address this limitation by proposing \method, a diffusion policy that jointly denoises bimanual actions and future 3D point tracks. By allowing action and track hypotheses to evolve together within a shared Transformer, each can inform and refine the other throughout denoising. We further ground multimodal representations in a shared spatiotemporal coordinate system to facilitate geometry-aware interaction during joint denoising. We evaluate \method on diverse bimanual manipulation tasks in RoboTwin 2.0 and on a real-world robot, comparing it with action-only policies and alternative future-prediction approaches spanning different state representations and learning objectives. Across 16 simulation tasks, \method achieves an average success rate of 83.4\%, outperforming the strongest baseline by 23.9 percentage points. On three real-world tasks, it outperforms the action-only and auxiliary geometry prediction methods by 50.0 and 21.2 percentage points, respectively. Beyond these performance gains, \method shows stronger generalization to cluttered scenes and out-of-distribution backgrounds than the evaluated baselines. Together, these results demonstrate the effectiveness of our joint action–motion modeling framework for coordinated bimanual manipulation. Our project website is available at \url{https://jam-bimanual.github.io/}.
\end{abstract}

\section{INTRODUCTION}

Bimanual manipulation requires two robot arms to coordinate their motions while interacting with shared objects and the surrounding environment~\cite{zhao2023aloha,grannen2023stabilize,lu2025anybimanual,wang2026cubic}. Although geometric reasoning is important for manipulation in general, it becomes particularly critical in bimanual settings, where the two arms are coupled through object contacts and coordinated interactions~\cite{liu2025voxactb,lv2025kstar}. The motion of one arm can alter the object pose and the feasible motion of the other, making successful execution depend on their joint effect on the scene. A bimanual policy must therefore reason not only about the immediate commands of each arm, but also about how their coordinated actions are expected to change the scene over time~\cite{chen2025stateprediction,xu2025imaginative,xu2026gap}.

Diffusion policies provide a strong framework for bimanual manipulation because they can generate temporally coherent action chunks and model complex action distributions~\cite{chi2023diffusion}. However, most diffusion policies are trained only to reconstruct demonstrated actions from the current observation. Their objectives do not explicitly connect action generation with predictions of the resulting object motion or scene changes, limiting the policy’s ability to coordinate the two arms around their effects on shared objects.

\begin{figure}[t]
    \centering
    \includegraphics[width=\linewidth]{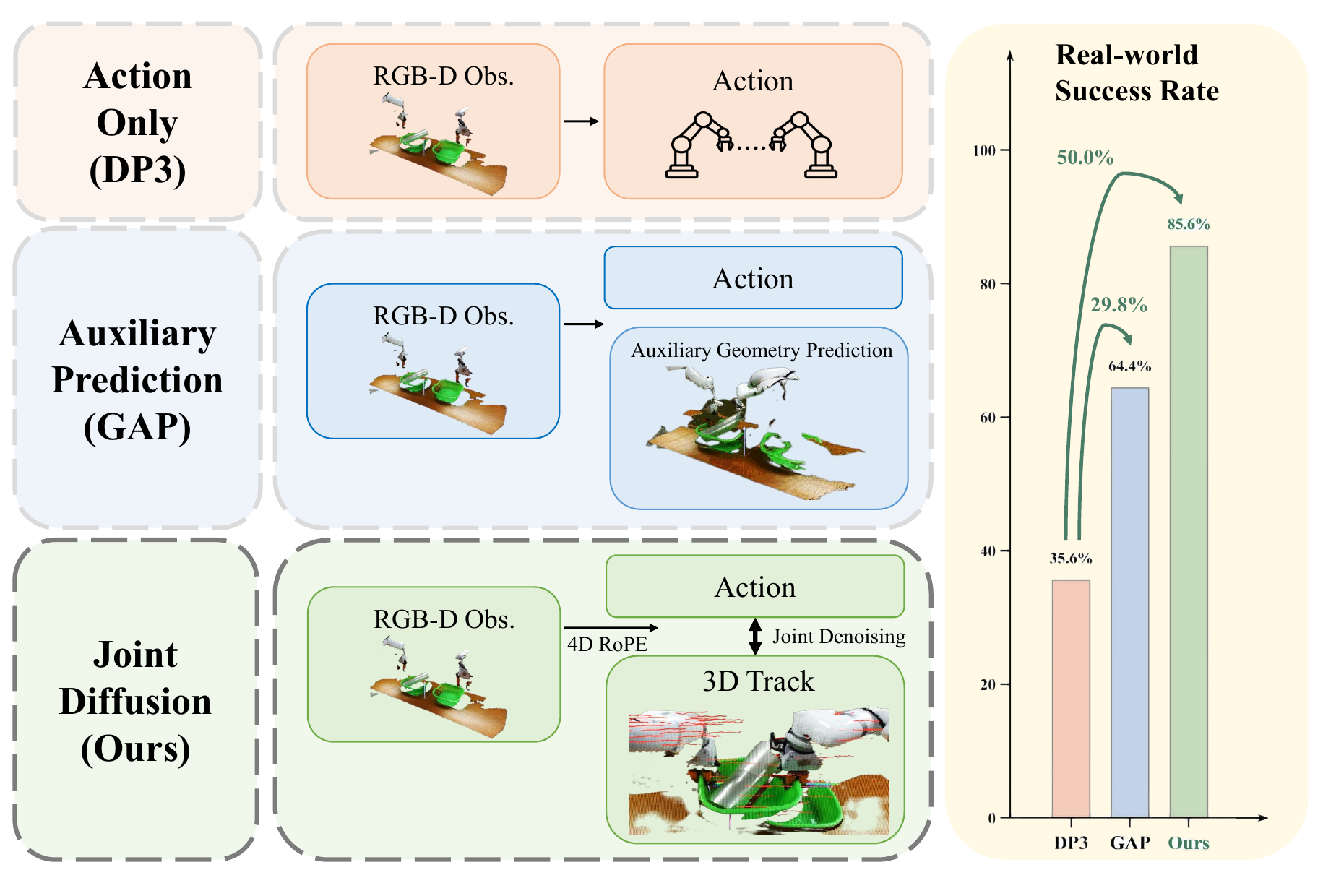}
    \caption{Paradigm comparison and real-world experiment results. Action-only method denoises robot actions, while auxiliary geometry prediction method augments action prediction with terminal 3D geometry feature prediction. In contrast, our method jointly denoises actions and future 3D tracks, enabling mutual refinement. Our method achieves an average success rate of \textbf{85.6\%}, compared with 64.4\% for GAP and 35.6\% for DP3.}
    \label{fig:teaser}
\end{figure}

Recent approaches address this limitation by predicting future images, video latents, or dense 3D geometry alongside robot actions~\cite{guo2024pad,xu2025imaginative,zhu2025uwm,xu2026gap,han2026geometricactionmodelrobot}. The choice of future representation determines what information is made available to the policy. Pixel- and video-based representations preserve rich visual context, but also model appearance details that may be irrelevant to control. Dense 3D geometry prediction provides rich spatial information, but predicting the complete scene often wastes computational capacity on representing task-irrelevant regions. For manipulation, it is more useful to explicitly model how scene elements move and evolve over time. Point tracks provide such a motion representation by preserving point correspondence and trajectory structure across time. Recent work shows that jointly predicting 2D tracks and actions can improve world--action modeling~\cite{guan2026jopat, cao2026tract}. Yet image-plane tracks do not directly encode metric depth or 3D displacement, which are important for reasoning about the relative geometry among two arms and manipulated scene objects. Moreover, track tokens possess explicit spatial and temporal relationships that may be difficult to recover from image features alone. These observations motivate us to investigate 3D point tracks with an explicit spatiotemporal positional structure as a future representation for bimanual control.

Our key idea is to couple future 3D motion prediction directly with bimanual action generation. Candidate actions determine how the scene is expected to evolve, while predicted point tracks provide a geometric description of the outcome that those actions should produce. We therefore jointly model bimanual actions and future 3D point tracks within a unified diffusion framework, allowing evolving action and motion predictions to mutually refine each other. To support this interaction, we ground multimodal tokens in a shared coordinate system defined by world-space positions and trajectory time, using 4D rotary positional encoding to incorporate their relative spatial and temporal relationships into attention.

We evaluate the approach on 16 simulated bimanual manipulation tasks in RoboTwin 2.0~\cite{chen2025robotwin2} and three tasks on a real-world bimanual robot system. \method outperforms the strongest baseline by 23.9 percentage points in simulation and achieves consistent improvements in the real world. \method also demonstrates stronger generalization to cluttered scenes and out-of-distribution backgrounds than the evaluated baselines. Ablation studies further demonstrate the benefits of joint action--motion denoising over alternative prediction designs and examine how spatiotemporal grounding affect policy performance.

Our contributions are summarized as follows:
\begin{itemize}
    \item We jointly denoise bimanual actions and 3D point tracks, allowing action and future-motion predictions to mutually refine each other.

    \item We ground multimodal representations in a shared spatiotemporal coordinate system, enabling geometry-aware interactions during joint action–motion denoising.

    \item We demonstrate strong performance on simulated and real-world bimanual tasks, along with improved generalization to out-of-distribution scenarios compared with the evaluated baselines.
\end{itemize}

\begin{figure*}[t]
    \centering
    \includegraphics[width=0.95\textwidth]{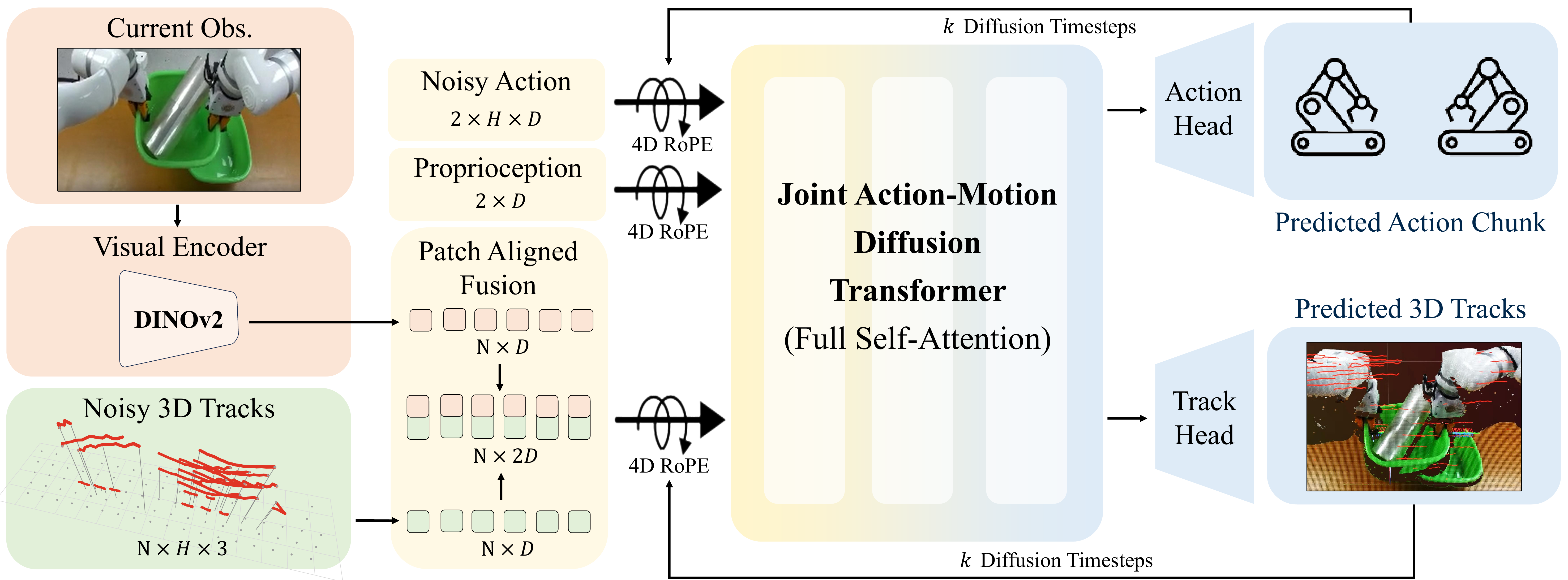}
    \caption{Overview of \method. Patch tokens from a DINOv2 encoder are fused with patch-aligned noisy 3D track queries and placed in a single DiT sequence with the arm-state tokens and the noisy bimanual action chunk. Full self-attention lets actions and future 3D tracks be denoised jointly and refine each other, while 4D rotary positional encoding grounds all tokens in a shared world-space and trajectory-time frame. Only the denoised actions are executed; the denoised 3D tracks act as an internal predictive representation and for visualization.}
    \label{fig:overview}
\end{figure*}

\section{Related Work}

\subsection{Bimanual Visuomotor Policy Learning}

Learning visuomotor policies from demonstrations has been widely studied through action-chunking and generative policy architectures, including ACT~\cite{zhao2023aloha}, Diffusion Policy~\cite{chi2023diffusion}, and its 3D extension DP3~\cite{ze20243d}. Building on these general policy-learning frameworks, recent work has explored different mechanisms for bimanual coordination, including modeling asymmetric acting and stabilizing roles between the two arms~\cite{grannen2023stabilize,liu2025voxactb}, transferring pretrained unimanual skills to coordinated dual-arm manipulation~\cite{lu2025anybimanual,im2026twinvla,bajamahal2026monoduo,song2026energyaction,franzese2023interactive}, and incorporating spatial and kinematic constraints into policy learning~\cite{lv2025kstar}. Complementary work grounds visual and action tokens in a common 3D coordinate frame, using relative spatial attention to guide action generation~\cite{gkanatsios20253d,ke20243d}.

These approaches primarily improve coordination at the action level, without explicitly modeling the future scene outcomes induced by those actions. Such outcomes are important for bimanual manipulation because coordinated actions must jointly produce the desired object motion and spatial configuration. Our work models these consequences through future 3D point tracks, jointly predicting them with bimanual actions.

\subsection{Future Prediction for Robot Policy}

Future prediction has been incorporated into robot policies through various modalities, including visual observations, geometric representations, and motion representations. Visual world-action models jointly predict future images~\cite{du2023learning,guo2024pad,xu2025imaginative,zhu2025uwm} or videos~\cite{li2025unifiedvideoactionmodel,ma2026dit4dit,motubrainteam2026motubrainadvancedworldaction,yuan2026fast,zhou2026zero} alongside robot actions, using future visual prediction during policy learning or as auxiliary supervision. While effective for modeling scene evolution, visual predictions do not explicitly encode metric 3D structure and may devote capacity to appearance details less relevant to control. Another line of work predicts geometric representations such as future depth maps~\cite{han2026geometricactionmodelrobot} and 3D geometry features~\cite{xu2026gap}, providing a more physically grounded description of future scene states. However, dense future prediction can be computationally expensive and may allocate substantial capacity to modeling static regions that are less informative for control; further, predicting future scene states does not explicitly capture how scene motion evolves over time.

Motion representations instead focus directly on scene dynamics. Prior work has modeled motion in image space using optical flow~\cite{ranasinghe2026futureopticalflowprediction} and point tracks, which preserve temporal point correspondence and have been used for action prediction and cross-embodiment transfer~\cite{collins2025amplify,wen2023any,bharadhwaj2024track2actpredictingpointtracks,xu2024flowcrossdomainmanipulationinterface}. More recent works further jointly generate point tracks and actions~\cite{guan2026jopat,cao2026tract}. However, these approaches represent tracks in 2D image space rather than in a shared 3D coordinate frame with robot end-effector motion. Existing methods also incorporate 3D motion into policy learning, using predicted motion to guide robot actions~\cite{wang2026lamplearningvisionlanguageactionpolicies,hung20263pointr3dpointtracks,lin2026chronoflowpolicyunifyingpastcurrentfutureinteraction,lee2026tracegen,tong2026pointaction, lin2026roboflow4d}, or using motion supervision and distilled tracker features to support action prediction~\cite{li2026egowam,wang2026track4actiondistillingworldcentric3d}. However, these methods do not jointly update explicit future motion and action throughout generation. In contrast, we treat future 3D point tracks and bimanual actions as diffusion variables that interact and are jointly refined at each denoising step. Each track query is further paired with its aligned visual patch, while spatiotemporal positional encoding grounds multimodal tokens in a shared 3D coordinate frame.

\section{PROBLEM FORMULATION}

We assume a dataset of expert trajectories
\begin{equation}
\mathcal{D}=\{\tau_i\}_{i=1}^{M},\qquad
\tau_i=\{(o_t,s_t,a_t)\}_{t=1}^{T_i},
\end{equation}
where $o_t$ denotes the visual observation, $s_t$ denotes the robot proprioceptive state, and $a_t\in\mathbb{R}^{D_a}$ denotes the bimanual robot action. At time step $t$, the policy predicts an action chunk
\begin{equation}
A_t=[a_t,\ldots,a_{t+H-1}]
\in\mathbb{R}^{H\times D_a}.
\end{equation}

To represent future scene motion, we associate the visual observation with a grid of $N$ patch-aligned 3D query points. Let $q_{i,0}\in\mathbb{R}^{3}$ denote the current position of query $i$, and let $q_{i,h}$ denote its position at future step $h$. We represent its future trajectory using displacements from the current position:
\begin{equation}
P_t(i,h)=q_{i,h}-q_{i,0},
\qquad
P_t\in\mathbb{R}^{N\times H\times 3}.
\label{eq:track_target}
\end{equation}
Each track query corresponds to a visual patch, providing aligned visual and motion representations.

Our objective is to learn the conditional joint distribution
\begin{equation}
p_\theta(A_t,P_t\mid o_t,s_t),
\end{equation}
so that the action and future-motion predictions can be jointly generated and refined.

\section{Method}

\subsection{3D Track Construction}

We construct patch-aligned 3D point tracks as an explicit representation of future scene motion. For each visual patch, we initialize one image-space query at the patch center, yielding a fixed grid of $N$ points distributed over the input image. Each query is then lifted into 3D using the corresponding depth and camera geometry, producing a set of scene points whose motion is tracked over the future horizon.

Let $q_{i,0}\in\mathbb{R}^{3}$ denote the current 3D position corresponding to query $i$, and let $q_{i,h}$ denote its position at future step $h$. We represent the trajectory using displacements from the current position according to Equation~\ref{eq:track_target}. Compared with pixel-dense motion representations, the patch-level query grid provides a lower-dimensional description of 3D scene motion while preserving point correspondence over time.

\subsection{Joint Action--Motion Diffusion}

We use a Diffusion Transformer (DiT)~\cite{peebles2022dit} to jointly model bimanual action sequences and future 3D tracks. A DINOv2 encoder~\cite{oquab2023dinov2} maps the current RGB observation to a set of patch tokens:
\begin{equation}
V=E_{\mathrm{O}}(o_t)
=[v_1,\ldots,v_N].
\end{equation}
The proprioceptive states of the left and right arms are encoded separately:
\begin{equation}
x_L^S=E_S(s_t^L),
\qquad
x_R^S=E_S(s_t^R).
\end{equation}

Let $A_t^k$ and $P_t^k$ denote the noisy action and track variables at diffusion step $k$. Because each track query corresponds to one visual patch, we concatenate its track embedding with the aligned visual feature and project them into a fused token:
\begin{equation}
z_i^k=
E_{\mathrm{fuse}}
\left(
v_i \Vert E_P(P_t^k(i,:))
\right),
\label{eq:visual_track_fusion}
\end{equation}
where $\Vert$ denotes feature concatenation. Each step of the noisy action chunk is encoded as:
\begin{equation}
x_j^{A,k}=E_A\big(a_{t+j-1}^k\big),
\qquad j=1,\ldots,H,
\label{eq:action_token}
\end{equation}
The complete DiT sequence consists of the fused visual--track tokens, two arm-state tokens, noisy bimanual action tokens, and a diffusion-timestep token:
\begin{equation}
X^k=
[z_{1:N}^k;
x_L^S;
x_R^S;
x_{1:H}^{A,k};
x_{\mathrm{diff}}^k].
\label{eq:dit_sequence}
\end{equation}
All tokens interact through full self-attention, allowing the action and future-motion hypotheses to mutually refine each other throughout denoising.

We combine learnable positional embeddings, which encode token identity and sequence role, with 4D rotary positional encoding (RoPE)~\cite{nakura2026empirical}, which encodes relative world-space positions and trajectory time. To establish this shared spatiotemporal frame, fused visual--track tokens are anchored at their current 3D patch positions and state tokens at the current end-effector positions, both with time index zero. Action tokens are assigned future-step indices and spatial positions derived from the noisy actions. Since these positions are unreliable at high noise levels,
we interpolate the action tokens' RoPE anchors between the current end-effector positions and the action-derived positions, gradually increasing the interpolation weight assigned to the action-derived positions as denoising progresses.

\subsection{Training and Inference}

Before diffusion, we apply per-dimension min--max normalization to map actions to $[-1,1]$ and normalize 3D track displacements. 
We add noise to the normalized action and track targets using a shared diffusion timestep:
\begin{align}
\widetilde{A}_t^k
&= \alpha_k \widetilde{A}_t^0 + \sigma_k \epsilon_A,\\
\widetilde{P}_t^k
&= \alpha_k \widetilde{P}_t^0 + \sigma_k \epsilon_P,
\end{align}
where $\epsilon_A$ and $\epsilon_P$ are independently sampled standard Gaussian noise. The shared DiT predicts the corresponding targets through separate output projections:
\begin{equation}
(\widehat{Y}_A,\widehat{Y}_P)
=
D_\theta(X^k),
\end{equation}
where $X^k$ contains the conditioning tokens and noisy action and track tokens, and $Y_A$ and $Y_P$ denote the diffusion prediction targets defined in the normalized spaces.

Because the patch-aligned query grid contains many nearly static background points, we follow PointWorld~\cite{huang2026pointworld} and assign each track a motion-dependent weight based on its final displacement in metric space:
\begin{equation}
w_{b,i}
=
w_{\min}
+
(1-w_{\min})
\sigma\left(
\kappa
\left(
\left\|P_{b,i,H}^0\right\|_2-\tau
\right)
\right),
\label{eq:movement_weight}
\end{equation}
where $\sigma(\cdot)$ denotes the sigmoid function, $\tau$ is the motion threshold, $\kappa$ controls the transition sharpness, and $w_{\min}$ specifies the minimum weight assigned to nearly static points. Here, $P^0$ denotes the unnormalized track displacement.

The motion-weighted track loss is
\begin{equation}
\mathcal{L}_{\mathrm{track}}
=
\frac{
\displaystyle
\sum_{b,i}
w_{b,i}
\sum_{h=1}^{H}\sum_{d=1}^{3}
\left(
\widehat{Y}_{P,b,i,h,d}
-
Y_{P,b,i,h,d}
\right)^2
}{
\displaystyle
3\sum_{b,i}w_{b,i}+\varepsilon
},
\label{eq:track_loss}
\end{equation}
where $b$ indexes training samples and $i$ indexes patch-aligned track queries. The loss averages over spatial coordinates, sums across the prediction horizon, and takes a motion-weighted average over all track queries.
The complete training objective is
\begin{equation}
\mathcal{L}
=
\mathcal{L}_{\mathrm{action}}
+
\lambda_{\mathrm{track}}
\mathcal{L}_{\mathrm{track}},
\label{eq:total_loss}
\end{equation}
where $\mathcal{L}_{\mathrm{action}}$ is the mean squared error between the predicted and target action outputs, and $\lambda_{\mathrm{track}}$ balances the two objectives.

At inference time, normalized actions and tracks are initialized independently from Gaussian noise:
\begin{equation}
\widetilde{A}_t^K\sim\mathcal{N}(0,I),
\qquad
\widetilde{P}_t^K\sim\mathcal{N}(0,I),
\end{equation}
and updated jointly using the same denoising schedule. After the final denoising step, both outputs are de-normalized using their respective training-set statistics. Only the predicted actions are executed, while the generated tracks are retained for analysis and visualization.

\section{Experiments}
Our experiments investigate three main questions: \textbf{Q1:} Does predicting future 3D point tracks improve bimanual policy performance? \textbf{Q2:} How do joint action--motion diffusion and spatiotemporal grounding affect policy performance? \textbf{Q3:} Does our method generalize better to unseen backgrounds and object appearances?
\subsection{Experimental Setup}
\paragraph{Simulation benchmark}
We evaluate on RoboTwin 2.0~\cite{chen2025robotwin2}, following the two bimanual task categories used in prior work: 8 Sync-bimanual tasks, and 8 Seq-coordinate tasks. Sync-bimanual tasks require simultaneous coordination, and Seq-coordinate tasks require sequential interaction between the two arms. Table \ref{tab:sync_bimanual} and \ref{tab:seq_coordinate} report the complete per-task results for the two categories.

\paragraph{Real-world evaluation}
We evaluate on three real-world bimanual manipulation tasks---Store Block, Stack Basin, and Place Duck Box---using two xArm manipulators and a fixed ZED Mini stereo camera mounted above the workspace. Demonstrations are collected via teleoperation using a Meta VR headset. The policy uses the same observation and action modalities as in simulation. The three tasks cover complementary forms of coordination, including sequential transfer, coordinated object transport, and simultaneous dual-arm interaction.

\begin{figure}[t]
    \centering
    \includegraphics[width=0.88\linewidth, trim={10 5 10 5}, clip]{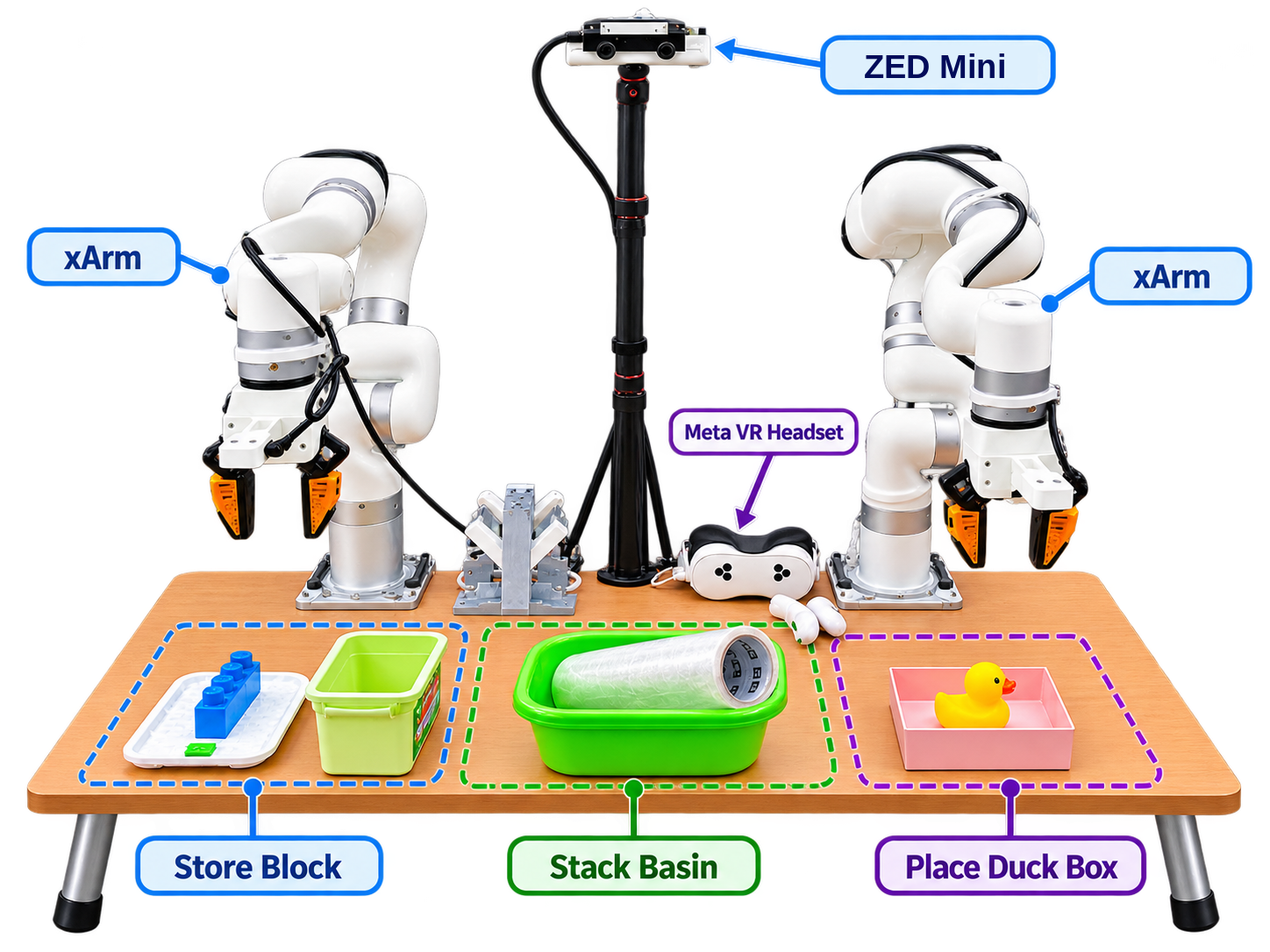}

    \vspace{-1mm}
    \caption{Real-world bimanual setup with two xArm manipulators, a fixed ZED Mini stereo camera, and a Meta VR headset for teleoperation. We evaluate on three tasks covering complementary forms of bimanual coordination.}
    \label{fig:real_tasks}
\end{figure}
\begin{table*}[t]
    \centering
    \caption{\textbf{Comparison on Sync-bimanual tasks (8 tasks).} Tasks requiring synchronized and coordinated operation of both arms. We report task success rate (\%) as mean $\pm$ standard deviation across three evaluation seeds. Best results are shown in \textbf{bold} and second-best results are \underline{underlined}.}
    \label{tab:sync_bimanual}
    \scriptsize
    \setlength{\tabcolsep}{1.85pt}
    \begin{tabular*}{\textwidth}{@{}l|@{\extracolsep{\fill}}ccccccccc@{}}
        \toprule
        & \textbf{\cellcolor{gray!15}Avg.$\uparrow$} & \textup{Pick Diverse} & \textup{Pick Dual} & \textup{Place Bread} & \textup{Place Bread} & \textup{Place Burger} & \textup{Place Cans} & \textup{Place Dual} & \textup{Put Bottles} \\
        \textbf{Method} & \cellcolor{gray!15}(\%) & \textup{Bottles} & \textup{Bottles} & \textup{Skillet} & \textup{Basket} & \textup{Fries} & \textup{Plastic Box} & \textup{Shoes} & \textup{Dustbin} \\
        \midrule
        DP~\cite{chi2023diffusion} & \cellcolor{gray!15} 45.8 & \result{38.0}{3.3} & \result{50.0}{8.6} & \result{38.7}{10.9} & \result{39.3}{0.9} & \result{\underline{97.3}}{0.9} & \result{68.0}{2.8} & \result{12.0}{4.3} & \result{23.3}{2.5} \\
        DP3~\cite{ze20243d} & \cellcolor{gray!15} 58.1 & \result{\underline{67.3}}{5.7} & \result{78.0}{3.3} & \result{49.3}{4.1} & \result{32.7}{5.0} & \result{85.3}{5.2} & \result{73.3}{5.7} & \result{0.0}{0.0} & \result{\underline{78.7}}{5.0} \\
        DICP~\cite{xu2025imaginative} & \cellcolor{gray!15} 57.5 & \result{23.3}{9.4} & \result{43.3}{3.4} & \result{45.3}{7.5} & \result{32.0}{7.1} & \result{96.7}{0.9} & \result{\underline{95.3}}{2.5} & \result{\underline{56.0}}{7.1} & \result{68.0}{1.6} \\
        GAP~\cite{xu2026gap} & \cellcolor{gray!15} 45.3 & \result{21.3}{6.2} & \result{76.0}{2.8} & \result{21.3}{5.0} & \result{16.0}{1.6} & \result{84.0}{4.3} & \result{40.0}{6.5} & \result{38.7}{7.4} & \result{65.3}{5.1} \\
         ATM~\cite{wen2023any} & \cellcolor{gray!15} 63.2 & \result{50.7}{6.8} & \result{\underline{78.7}}{1.9} & \result{\underline{50.7}}{3.8} & \result{\underline{58.0}}{3.3} & \result{96.7}{1.9} & \result{94.0}{4.3} & \result{14.7}{5.2} & \result{62.0}{2.8} \\
        \textbf{\method} & \cellcolor{gray!15} \textbf{85.2} & \bestresult{85.3}{0.9} & \bestresult{99.3}{0.9} & \bestresult{72.7}{5.1} & \bestresult{77.3}{1.9} & \bestresult{98.0}{1.6} & \bestresult{100.0}{0.0} & \bestresult{62.0}{5.9} & \bestresult{86.7}{0.9} \\
        \bottomrule
    \end{tabular*}
\end{table*}
\begin{table*}[t]
    \centering
    \caption{\textbf{Comparison on Seq-coordinate tasks (8 tasks).} Tasks requiring multi-step manipulation and long-horizon reasoning. We report task success rate (\%) as mean $\pm$ standard deviation across three evaluation seeds. Best results are shown in \textbf{bold} and second-best results are \underline{underlined}.}
    \label{tab:seq_coordinate}
    \scriptsize
    \setlength{\tabcolsep}{1.85pt}
    \begin{tabular*}{\textwidth}{@{}l|@{\extracolsep{\fill}}ccccccccc@{}}
        \toprule
        & \textbf{\cellcolor{gray!15}Avg.$\uparrow$} & \textup{Handover} & \textup{Hang} & \textup{Place Object} & \textup{Scan} & \textup{Stack} & \textup{Stack} & \textup{Stack} & \textup{Stack} \\
        \textbf{Method} & \cellcolor{gray!15}(\%) & \textup{Block} & \textup{Mug} & \textup{Basket} & \textup{Object} & \textup{Blocks Two} & \textup{Blocks Three} & \textup{Bowls Two} & \textup{Bowls Three} \\
        \midrule
        DP~\cite{chi2023diffusion} & \cellcolor{gray!15} 39.0 & \result{64.0}{6.5} & \result{20.7}{5.2} & \result{34.7}{6.6} & \result{20.7}{5.2} & \result{10.7}{1.9} & \result{0.0}{0.0} & \result{87.3}{1.9} & \result{74.0}{4.3} \\
        DP3~\cite{ze20243d} & \cellcolor{gray!15} 51.9 & \result{\underline{95.3}}{1.9} & \result{22.0}{1.6} & \result{52.0}{4.3} & \result{42.7}{5.2} & \result{38.0}{2.8} & \result{2.0}{1.6} & \result{92.7}{2.5} & \result{70.7}{1.9} \\
        DICP~\cite{xu2025imaginative} & \cellcolor{gray!15} 61.5 & \result{84.0}{3.3} & \result{\underline{37.3}}{4.7} & \result{\underline{62.0}}{4.3} & \result{27.3}{6.6} & \result{\underline{74.0}}{1.6} & \result{\underline{38.0}}{0.0} & \bestresult{94.7}{3.4} & \result{75.0}{5.0} \\
        GAP~\cite{xu2026gap} & \cellcolor{gray!15} 51.1 & \result{73.3}{1.9} & \result{24.7}{9.0} & \result{52.7}{4.7} & \result{\underline{45.0}}{3.0} & \result{41.3}{0.9} & \result{0.7}{0.9} & \result{92.0}{2.8} & \bestresult{78.7}{5.7} \\
        ATM~\cite{wen2023any} & \cellcolor{gray!15} 40.8 & \result{56.7}{2.5} & \result{24.7}{1.9} & \result{40.0}{7.1} & \result{28.7}{5.2} & \result{24.0}{1.6} & \result{8.0}{1.6} & \result{76.7}{5.0} & \result{67.3}{3.4} \\
        \textbf{\method} & \cellcolor{gray!15} \textbf{81.6} & \bestresult{96.7}{0.9} & \bestresult{54.0}{3.3} & \bestresult{70.7}{5.7} & \bestresult{72.0}{2.8} & \bestresult{91.3}{2.5} & \bestresult{96.0}{1.6} & \result{\underline{94.0}}{4.3} & \result{\underline{78.0}}{4.3} \\
        \bottomrule
    \end{tabular*}
\end{table*}

\paragraph{Observations and track supervision}
All policies receive a single fixed RGB view together with robot proprioception. In simulation, we construct ground-truth 3D tracks by projecting each current visual patch center onto the corresponding scene geometry and tracking the associated 3D point over the future horizon in the simulator. For real-world training, we first obtain 2D point tracks using CoTracker3~\cite{karaev2024cotracker3}. We estimate metric depth from rectified stereo RGB with FoundationStereo~\cite{wen2025foundationstereo} and use the estimated depth together with calibrated camera parameters to lift the tracks into 3D in a common world coordinate frame. At inference time, future observations and 3D tracks are not available to the policy. We estimate the current depth map using FastFoundationStereo~\cite{wen2026fast} and use it to recover the 3D positions required by spatiotemporal RoPE.

\paragraph{Compared methods}

We compare against baselines spanning both action-only policy learning and different forms of future prediction. \textbf{Action-only policies} include DP~\cite{chi2023diffusion} and DP3~\cite{ze20243d}, which directly predict action chunks from 2D and 3D observations, respectively. \textbf{Future-state prediction methods} augment action learning with predictions of future scene states: DICP~\cite{xu2025imaginative} predicts future visual representations, while GAP~\cite{xu2026gap} predicts dense geometric features at the endpoint alongside actions. \textbf{Trajectory-guided policies} are represented by ATM~\cite{wen2023any}, which first predicts future 2D point tracks and then conditions action prediction on the predicted tracks. In contrast, our method jointly denoises bimanual actions and full-horizon 3D point tracks, allowing the two predictions to refine each other throughout generation. All methods are trained and evaluated with matched demonstrations, observation modalities, action representations, and initial-state distributions whenever applicable. For real-world evaluation, we compare against DP3~\cite{ze20243d} and GAP~\cite{xu2026gap}.

\paragraph{Evaluation protocol}
In simulation, each policy is trained on 100 expert demonstrations per task and evaluated with three independent seeds, each over 50 rollouts with matched initial states. We report the mean and standard deviation of task success rate across seeds. In the real world, each policy is trained on 50 demonstrations per task and evaluated over 30 rollouts per method--task pair.

\paragraph{Implementation details}
For all experiments, the policy receives a single head-camera RGB observation with an observation horizon of $H_o=1$. We use an action horizon of $H_a=16$ and a track horizon of $H_p=16$. Images are resized to $238\times322$ and encoded with DINOv2~\cite{oquab2023dinov2} using a patch size of $14$, yielding a $17\times23$ patch grid and $N=391$ track queries, one per visual patch. The DiT consists of $4$ Transformer blocks with hidden dimension $1024$ and $8$ attention heads. We train with AdamW using a learning rate of $10^{-4}$, $(\beta_1,\beta_2)=(0.9,0.999)$, weight decay $10^{-6}$, cosine decay with $500$ warm-up steps, batch size $64$, and $200$ epochs. We use $100$ diffusion timesteps during training and $10$ DDIM denoising steps at inference. Real-world experiments use the same model and optimization settings, with regions outside the tabletop workspace cropped from the visual observations.

\subsection{Simulation Policy Results}
\begin{figure}[h]
    \centering
    \includegraphics[width=0.95\linewidth]{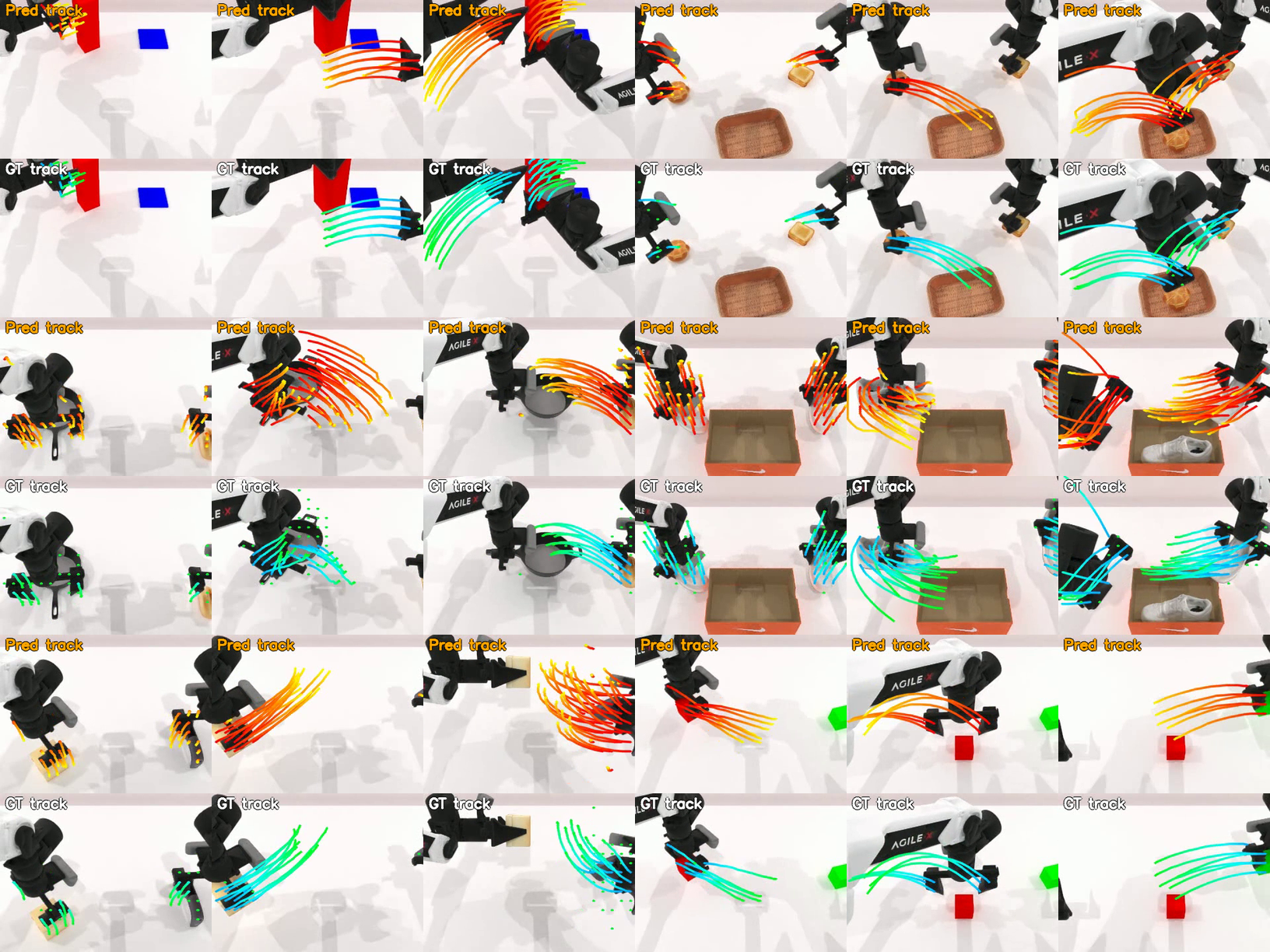}
    \caption{Qualitative visualization of future 3D point tracks predicted by our method and the corresponding ground-truth tracks on RoboTwin. Red--yellow tracks denote predictions, while green--cyan tracks denote ground truth.}
    \label{fig:track_visualization}
\end{figure}

Tables~\ref{tab:sync_bimanual} and~\ref{tab:seq_coordinate} report per-task success rates across the two RoboTwin task categories. \method achieves an average success rate of $85.2\%$ on Sync-bimanual tasks, which require simultaneous coordination, and $81.6\%$ on Seq-coordinate tasks, which require sequential interaction between the two arms. Across all 16 tasks, \method achieves an overall success rate of $83.4\%$, outperforming the strongest baseline by $23.9$ percentage points.
Fig.~\ref{fig:track_visualization} shows that 3D point tracks predicted by \method closely match ground truth in simulation.

The results show a consistent benefit from explicitly modeling future motion. Compared with action-only policies such as DP~\cite{chi2023diffusion} and DP3~\cite{ze20243d}, \method provides the policy with an additional representation of how the observed scene is expected to evolve.

We further compare against methods that incorporate future information. DICP~\cite{xu2025imaginative} predicts future visual observations, while GAP~\cite{xu2026gap} predicts future 3D geometry at the prediction horizon. In contrast, \method preserves the motion between the current and future states through temporally aligned 3D point tracks. Its stronger performance on coordination-intensive and multi-stage tasks suggests that this intermediate motion structure provides useful guidance that is not captured by either future visual observation or geometric representation alone.

\subsection{Real-World Bimanual Evaluation}
\begin{figure}[h]
    \centering
    \includegraphics[width=0.95\linewidth]{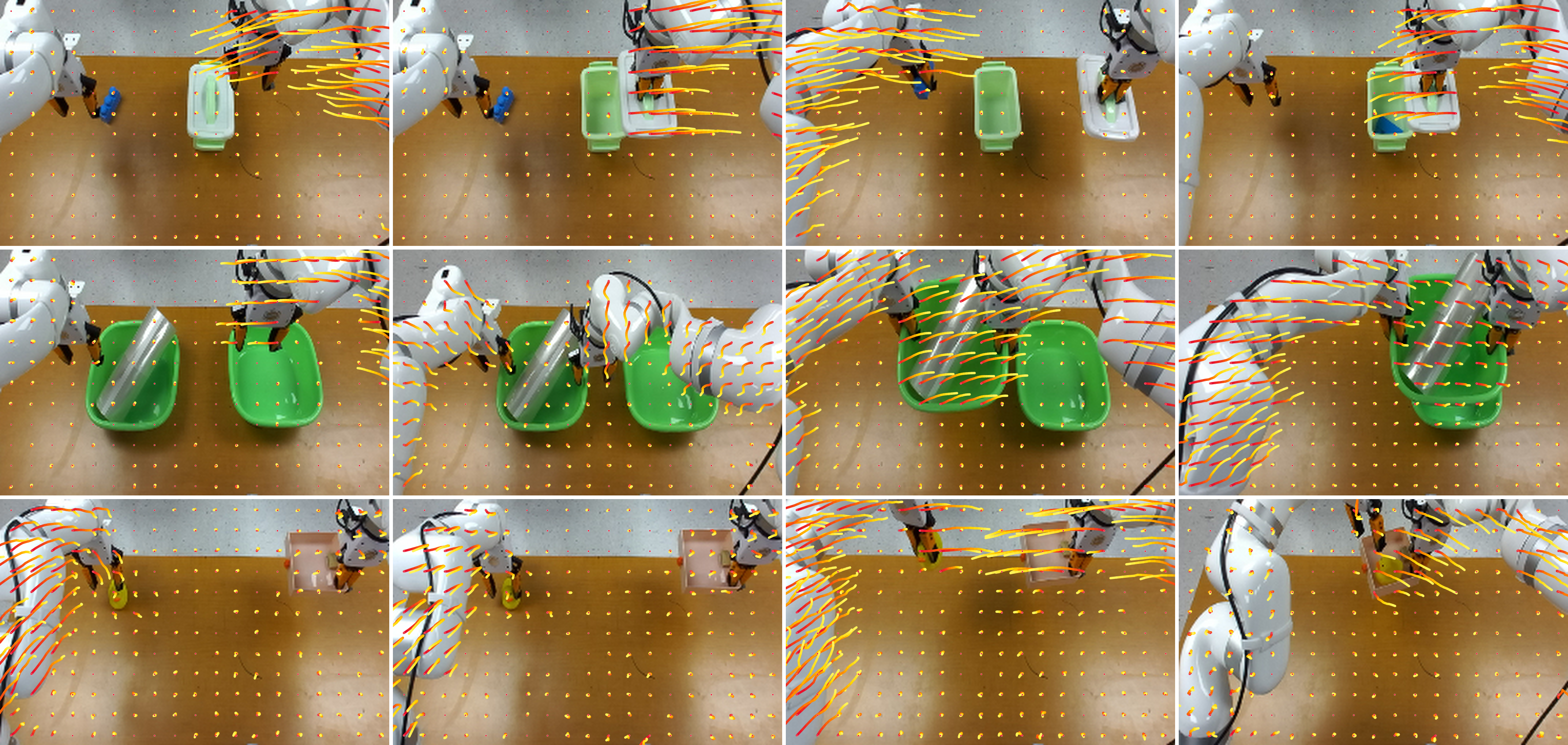}
    \caption{Qualitative visualization of future 3D point tracks predicted by our method on three real-world tasks: Store Block (top), Stack Basin (middle), and Place Duck Box (bottom). Predicted future 3D point tracks are color-coded by time, progressing from red to yellow.}
    \label{fig:real_task_sequences}
\end{figure}
Table~\ref{tab:real_results} reports the real-world evaluation. \method achieves the best performance on all three tasks, averaging $85.6\%$ and outperforming GAP~\cite{xu2026gap} by $21.2$ percentage points. The gap is small on \textit{Stack Basin}, where static future geometry already provides a strong cue, but grows substantially on \textit{Place Duck Box} ($80.0\%$ vs.\ $33.3\%$), suggesting that temporally structured tracks are more useful when success depends on coordinated motion throughout execution. \method also consistently outperforms DP3~\cite{ze20243d} without directly encoding dense point clouds, instead incorporating 3D geometry through positional encoding and explicit future-motion prediction.
Fig.~\ref{fig:real_task_sequences} shows that \method predicts coherent task-relevant future motion in real-world manipulation.

\begin{table}[t]
\caption{\textbf{Real-world task success rate (\%).} Each method is evaluated over 30 trials per task. Best results are shown in \textbf{bold}.}
\label{tab:real_results}
\centering
\small
\setlength{\tabcolsep}{4pt}
\resizebox{\columnwidth}{!}{
\begin{tabular}{lccccc}
\toprule
Method & \cellcolor{gray!15} Avg.$\uparrow$ & Store Block & Stack Basin & Place Duck Box \\
\midrule
DP3~\cite{ze20243d} & \cellcolor{gray!15} 35.6 & 23.3 & 76.7 & 6.7 \\
GAP~\cite{xu2026gap} & \cellcolor{gray!15} 64.4 & 73.3 & 86.7 & 33.3 \\
\textbf{\method} & \textbf{\cellcolor{gray!15} 85.6} & \textbf{83.3} & \textbf{93.3} & \textbf{80.0} \\
\bottomrule
\end{tabular}}
\end{table}

\subsection{Ablation Study}
\label{sec:ablation} We conduct controlled ablations on eight representative RoboTwin tasks, including four Sync-bimanual tasks and four Seq-coordinate tasks. All variants use the same visual encoder, action representation, training data, and DiT capacity whenever possible.

\paragraph{Future-track modeling}
We first study how future-track prediction should interact with action generation. Results can be found in Table~\ref{tab:architecture_ablation}. \emph{Action-only} removes track prediction entirely, while \emph{Track regression} retains visual--track fusion and shared self-attention but predicts clean tracks without maintaining an evolving track diffusion state. This variant raises average success from $61.4\%$ to $73.9\%$, supporting the value of future-motion supervision. \emph{Track-conditioned action diffusion} first predicts future tracks and uses this fixed prediction to condition action denoising, reaching $76.3\%$. The full model achieves $81.7\%$ by jointly denoising actions and tracks. Its advantage over these alternatives is evident on both synchronized and sequential tasks, suggesting that joint refinement benefits both simultaneous and multi-stage coordination. Removing 4D RoPE reduces average policy success from $81.7\%$ to $74.1\%$, suggesting that explicit spatiotemporal grounding across modalities benefits policy performance.

\begin{table}[h]
\caption{\textbf{Ablation study} on eight RoboTwin tasks. We report average success rates (\%) over three evaluation seeds, with 50 rollouts per seed per task. Best results are shown in \textbf{bold}.}
\label{tab:architecture_ablation}
\centering
\small
\setlength{\tabcolsep}{5pt}
\begin{tabular}{lccc}
\toprule
Variant & \cellcolor{gray!15} Avg.$\uparrow$ & Sync. & Seq. \\
\midrule
Action-only
& \cellcolor{gray!15} 61.4 & 57.5 & 65.3 \\

Track regression
& \cellcolor{gray!15} 73.9 & 68.4 & 79.3 \\

Track-conditioned action diffusion
& \cellcolor{gray!15} 76.3 & 72.8 & 79.8 \\


w/o 4D RoPE
& \cellcolor{gray!15} 74.1 & 65.7 & 82.5 \\

\textbf{Full model}
& \cellcolor{gray!15} \textbf{81.7} & \textbf{74.3} & \textbf{89.0} \\
\bottomrule
\end{tabular}
\end{table}


\begin{table}[t]
\caption{\textbf{Future 3D track prediction} on the validation sets of the eight RoboTwin tasks. We report ADE and FDE averaged across tasks. Best results are shown in \textbf{bold}.}
\label{tab:track_prediction}
\centering
\small
\setlength{\tabcolsep}{3pt}
\begin{tabular}{@{}lcc@{}}
\toprule
Variant & ADE (mm) $\downarrow$ & FDE (mm) $\downarrow$ \\
\midrule
Track regression
& 11.1 & 14.8 \\
Track-conditioned action diffusion
& 9.9 & 12.5 \\
w/o 4D RoPE
& 13.2 & 17.2 \\
\textbf{Full model}
& \textbf{7.6} & \textbf{10.0} \\
\bottomrule
\end{tabular}
\end{table}

\paragraph{Track Prediction and Policy Performance}
We further examine the relationship between future-track accuracy
and manipulation success. Table~\ref{tab:track_prediction} reports
average displacement error (ADE) and final displacement error (FDE)
for moving patches on the validation sets of the eight ablation tasks.
The full model achieves the lowest ADE/FDE of $7.6$/$10.0$ mm
and the highest average success rate of $81.7\%$.
Compared with track-conditioned action diffusion, joint denoising
reduces ADE/FDE from $9.9$/$12.5$ mm and improves success
by $5.4$ percentage points.
Removing 
4D RoPE
increases prediction errors and reduces policy success,
supporting the benefits of 
shared spatiotemporal grounding.


\paragraph{Future representations and generalization to harder visual conditions}
We evaluate policy generalization by training all methods on
the \emph{Easy} setting and directly testing on the corresponding
\emph{Hard} setting across the eight ablation tasks, without
fine-tuning. Hard scenes introduce unseen object textures
and cluttered backgrounds.
As shown in Table~\ref{tab:easy_to_hard_generalization},
\method achieves the highest average success rate of $17.9\%$,
compared with $4.3\%$ for the strongest baseline GAP.
We further examine alternative future representations within
our framework in Table~\ref{tab:representation_generalization_avg}, by performing ablations to our policy architecture.
Our full-horizon 3D track formulation achieves higher success
than full-horizon 2D tracks ($9.3\%$) and
endpoint 3D geometry ($10.3\%$).
These results complement the baseline comparisons and support
the effectiveness of our 3D track-based formulation for
generalization to harder visual conditions.

\begin{table}[h]
\caption{\textbf{Easy-to-hard generalization} on eight RoboTwin tasks without fine-tuning. We report average success rates (\%) over three evaluation seeds, with 50 rollouts per seed per task. Best results are shown in \textbf{bold}.}
\label{tab:easy_to_hard_generalization}
\centering
\small
\setlength{\tabcolsep}{4pt}
\resizebox{\columnwidth}{!}{%
\begin{tabular}{lccccc}
\toprule
Method
& DP3~\cite{ze20243d}
& DICP~\cite{xu2025imaginative}
& GAP~\cite{xu2026gap}
& ATM~\cite{wen2023any}
& \textbf{\method} \\
\midrule
Avg. (\%)
& 1.7
& 3.9
& 4.3
& 0.0
& \textbf{17.9} \\
\bottomrule
\end{tabular}%
}
\end{table}

\begin{table}[t]
    \centering
    \caption{\textbf{Easy-to-hard generalization with alternative future representations} on eight RoboTwin tasks. We report average success rates (\%) over three evaluation seeds, with 50 rollouts per seed per task. Best results are shown in \textbf{bold}.}
    \label{tab:representation_generalization_avg}
    \small
    \setlength{\tabcolsep}{4pt}
    \begin{tabular}{lccc}
        \toprule
        Future prediction target
        & \cellcolor{gray!15}Avg.$\uparrow$
        & Sync. & Seq. \\
        \midrule
        2D tracks (full horizon)
        & \cellcolor{gray!15}9.3
        & 7.5 & 11.2 \\

        3D geometry (endpoint)
        & \cellcolor{gray!15}10.3
        & 14.2 & 6.5 \\

        \textbf{3D tracks (full horizon, ours)}
        & \cellcolor{gray!15}\textbf{17.9}
        & \textbf{21.3} & \textbf{14.5} \\
        \bottomrule
    \end{tabular}
\end{table}

\section{Limitations and Future Work}
Our method currently uses a single camera view and a fixed grid of
track queries, while real-world track supervision depends on the
accuracy of point tracking and depth estimation. Future work will explore
efficient multi-view fusion, adaptive query selection, and robustness
to noisy supervision. 
Incorporating
force or contact information is another direction for future work.

\section{CONCLUSION}

We presented \method, a diffusion policy that jointly generates robot actions and future 3D tracks in a shared self-attention sequence. By jointly denoising actions and tracks, the model allows future-motion and action hypotheses to refine one another rather than treating future prediction as an auxiliary objective or fixed condition. Experiments on RoboTwin and real-world bimanual manipulation show consistent improvements over action-only and future-aware baselines, with particularly strong gains on coordination-intensive and multi-stage tasks. Our ablations further highlight the importance of joint action--motion generation and spatiotemporal positional encoding. Together, these results support future 3D tracks as a practical intermediate representation for coordinated robot action generation.


\bibliographystyle{IEEEtran}
\bibliography{references}

\clearpage
\appendix

\subsection{Additional Experimental Results}
\label{app:additional_results}
We additionally evaluate \method on the Dominant-select task category from the RoboTwin 2.0 benchmark. Table~\ref{tab:dominant_select} complements our main bimanual evaluation by demonstrating \method's effectiveness on tasks that require selecting the appropriate arm.
\begin{table*}[t]
    \centering
    \caption{\textbf{Comparison on Dominant-select tasks (16 tasks).} Tasks requiring appropriate arm selection. We report task success rate (\%) as mean $\pm$ standard deviation across three evaluation seeds. Best results are shown in \textbf{bold} and second-best results are \underline{underlined}.}
    \label{tab:dominant_select}
    \scriptsize
    \setlength{\tabcolsep}{2.0pt}
    \begin{tabular*}{\textwidth}{@{}l@{\extracolsep{\fill}}cccccccc@{}}
        \toprule
        & \textbf{\cellcolor{gray!15}Avg.$\uparrow$} & \textup{Beat Block} & \textup{Place} & \textup{Move} & \textup{Open} & \textup{Open} & \textup{Place A2B} & \textup{Place A2B} \\
        \textbf{Method} & \cellcolor{gray!15}(\%) & \textup{Hammer} & \textup{Shoe} & \textup{Can Pot} & \textup{Laptop} & \textup{Microwave} & \textup{Left} & \textup{Right} \\
        \midrule
        DP~\cite{chi2023diffusion} & \cellcolor{gray!15} 41.7 & \bestresult{92.7}{5.0} & \result{46.0}{7.1} & \result{70.0}{15.6} & \result{64.0}{4.3} & \result{0.0}{0.0} & \result{17.3}{4.7} & \result{17.3}{2.5} \\
        DP3~\cite{ze20243d} & \cellcolor{gray!15} 64.4 & \result{\underline{88.0}}{1.6} & \result{62.0}{2.8} & \bestresult{92.7}{1.9} & \result{39.3}{1.9} & \result{22.0}{1.6} & \bestresult{53.3}{0.9} & \bestresult{50.7}{5.7} \\
        DICP~\cite{xu2025imaginative} & \cellcolor{gray!15} 48.3 & \result{86.0}{3.3} & \result{\underline{72.0}}{7.1} & \result{74.7}{3.4} & \result{80.0}{4.3} & \result{16.0}{4.3} & \result{8.0}{1.6} & \result{10.0}{1.6} \\
        GAP~\cite{xu2026gap} & \cellcolor{gray!15} 51.7 & \result{39.3}{1.9} & \result{60.7}{2.5} & \result{66.0}{8.8} & \result{\underline{81.3}}{2.5} & \result{\underline{36.0}}{8.0} & \result{14.0}{1.6} & \result{11.3}{3.4} \\
        ATM~\cite{wen2023any} & \cellcolor{gray!15} 47.5 & \result{84.7}{2.5} & \result{40.7}{5.7} & \result{\underline{91.3}}{3.8} & \result{68.7}{3.8} & \result{1.3}{1.9} & \result{\underline{24.7}}{2.5} & \result{14.7}{0.9} \\
        \textbf{\method} & \cellcolor{gray!15} \textbf{73.5} & \result{86.7}{6.2} & \bestresult{96.0}{1.6} & \result{90.0}{2.8} & \bestresult{92.7}{0.9} & \bestresult{83.3}{2.5} & \result{23.3}{4.1} & \result{\underline{24.0}}{2.8} \\
        \midrule
        \multicolumn{1}{c}{\textup{Turn}} & \textup{Place} & \textup{Place Container} & \textup{Press} & \textup{Place Phone} & \textup{Rotate} & \textup{Place} & \textup{Place Object} & \textup{Shake} \\
        \multicolumn{1}{c}{\textup{Switch}} & \textup{Fan} & \textup{Plate} & \textup{Stapler} & \textup{Stand} & \textup{QR Code} & \textup{Empty Cup} & \textup{Stand} & \textup{Bottle} \\
        \midrule
        \multicolumn{1}{c}{\result{17.3}{3.4}} & \result{4.7}{4.1} & \result{51.3}{3.4} & \result{20.0}{1.6} & \result{38.0}{4.9} & \result{24.0}{1.6} & \result{61.3}{4.1} & \result{48.7}{1.9} & \result{94.7}{1.9} \\
        \multicolumn{1}{c}{\result{\underline{36.7}}{6.6}} & \result{\underline{50.7}}{5.0} & \result{\underline{92.0}}{1.6} & \result{67.3}{2.5} & \result{54.0}{4.9} & \bestresult{69.3}{2.5} & \result{\underline{75.3}}{8.2} & \bestresult{76.7}{3.4} & \bestresult{100.0}{0.0} \\
        \multicolumn{1}{c}{\result{25.3}{3.8}} & \result{40.0}{4.3}  & \result{86.7}{4.7} & \result{39.3}{0.9} & \result{11.3}{5.7} & \result{50.7}{1.9} & \result{52.0}{3.3} & \result{36.0}{4.3} & \result{84.0}{1.6}\\
        \multicolumn{1}{c}{\result{34.7}{1.9}} & \result{45.3}{4.7} & \result{78.0}{9.1} & \result{\underline{68.0}}{3.3} & \result{30.0}{8.6} & \result{62.0}{4.3} & \result{61.3}{6.6} & \result{39.3}{4.1} & \result{\underline{99.3}}{0.9} \\
        \multicolumn{1}{c}{\result{23.3}{2.5}} & \result{4.0}{0.0} & \result{69.3}{8.1} & \result{31.3}{6.2} & \result{\underline{58.0}}{4.9} & \result{48.0}{4.3} & \result{64.7}{8.1} & \result{40.7}{6.6} & \result{94.0}{4.3} \\
        \multicolumn{1}{c}{\bestresult{38.7}{1.9}} & \bestresult{62.7}{1.9} & \bestresult{98.0}{1.6} & \bestresult{80.7}{7.7} & \bestresult{90.0}{2.8} & \result{\underline{62.0}}{2.8} & \bestresult{94.7}{0.9} & \result{\underline{54.7}}{8.4} & \result{98.7}{0.9} \\
        \bottomrule
    \end{tabular*}
\end{table*}

\subsection{Future Track Prediction Metrics}
\label{app:track_metrics}
We evaluate future 3D track prediction using average displacement
error (ADE) and final displacement error (FDE).
Both metrics are computed over moving patches, defined as
queries whose ground-truth endpoint displacement satisfies
$\|P_t(i,H_p)\|_2 > 0.01\,\mathrm{m}$.
For each validation sample, we re-index these queries as
$i=1,\ldots,N_{\mathrm{mov}}$ and define
\begin{align}
\mathrm{ADE}
&=
\frac{1}{N_{\mathrm{mov}}H_p}
\sum_{i=1}^{N_{\mathrm{mov}}}
\sum_{h=1}^{H_p}
\left\|
\widehat{P}_t(i,h)-P_t(i,h)
\right\|_2,
\label{eq:track_ade}\\
\mathrm{FDE}
&=
\frac{1}{N_{\mathrm{mov}}}
\sum_{i=1}^{N_{\mathrm{mov}}}
\left\|
\widehat{P}_t(i,H_p)-P_t(i,H_p)
\right\|_2,
\label{eq:track_fde}
\end{align}
where $N_{\mathrm{mov}}$ is the number of selected moving
queries, $H_p$ is the prediction horizon, and
$\widehat{P}_t(i,h)$ and $P_t(i,h)$ denote the predicted
and ground-truth 3D displacements relative to the query's
anchor position, respectively.
ADE averages the Euclidean displacement error across the
full prediction horizon, whereas FDE measures the error
at the final step. Both metrics use de-normalized
displacements in meters, converted to millimeters for
reporting. Lower values indicate better prediction accuracy.

\subsection{Ablation Settings and Per-Task Results}
\label{app:ablation_results}
We construct ablation variants from the full model to examine
future-track supervision, action--track coupling, visual--track
fusion, and spatiotemporal grounding. Unless otherwise specified,
we retain the remaining model components and training settings.
\paragraph{Ablation settings}
The variants are implemented as follows:
\begin{itemize}
    \item \textbf{Action-only.}
    This variant retains the action-generation backbone but
    removes the track-prediction branch and all 3D track
    supervision.

    \item \textbf{Track regression.}
    We replace track diffusion with deterministic regression
from fixed track queries. Each query remains concatenated
with its corresponding visual feature and participates
in shared self-attention with action tokens.
The regression head predicts clean tracks at every
action-denoising step, but these predictions are not
fed back as an evolving track diffusion state.
We retain the prediction from the final denoiser forward pass.

    \item \textbf{Track-conditioned action diffusion.}
    This variant uses two-stage prediction. In the first stage,
    noisy 3D track tokens are concatenated with their corresponding
    patch-level visual features and denoised to predict future
    3D tracks. In the second stage, the predicted tracks are
    concatenated with the corresponding visual features and
    participate in self-attention with action tokens during
    action denoising. The track predictions remain fixed
    throughout the second stage.

    \item \textbf{w/o Visual--Track Concatenation.}
    We remove the concatenation of each track representation
    with its corresponding patch-level visual feature.
    Instead, track and action tokens interact with separate
    visual tokens through cross-attention. The model must
    therefore learn visual--track associations through
    attention rather than receiving explicitly paired features.

    \item \textbf{w/o 4D RoPE.}
    We remove the rotary encoding of spatial coordinates and
    trajectory time while retaining learnable positional
    embeddings for token identity and sequence role.
    All other components of joint action--track denoising
    remain unchanged.
\end{itemize}

\paragraph{Evaluation protocol}
Table~\ref{tab:per_task_ablation} reports success rates on the
eight RoboTwin tasks used for ablation. For each variant and
task, we evaluate the trained policy using three evaluation
seeds, with 50 rollouts per seed, and report the mean success
rate across seeds.
Table~\ref{tab:track_prediction_appendix} reports average
displacement error (ADE) and final displacement error (FDE)
for moving patches on the corresponding validation sets.
Both metrics are computed using de-normalized 3D displacements,
reported in millimeters, and averaged across tasks.

\paragraph{Track prediction and policy performance}
The full model achieves the lowest ADE/FDE of $7.6$/$10.0$ mm
and the highest average policy success rate of $81.7\%$.
Compared with track-conditioned action diffusion,
the full model reduces ADE/FDE from $9.9$/$12.5$ mm
and improves success by $5.4$ percentage points.
Removing 4D RoPE increases ADE/FDE to $13.2$/$17.2$ mm
and reduces success to $74.1\%$, supporting the benefits
of shared spatiotemporal grounding for both prediction
and control.
Removing visual--track concatenation also increases
ADE/FDE to $11.9$/$15.3$ mm and reduces success to $70.0\%$.
This comparison supports the effectiveness of the full
model's patch-aligned fusion design relative to the
separate-token cross-attention variant.

However, lower track error does not consistently translate
into higher policy success. Track regression predicts
more accurate tracks than the variant without 4D RoPE
($11.1$/$14.8$ mm vs.~$13.2$/$17.2$ mm), yet their success
rates are similar ($73.9\%$ vs.~$74.1\%$).
Likewise, the variant without visual--track concatenation
has lower track errors than the variant without 4D RoPE,
but a lower success rate ($70.0\%$ vs.~$74.1\%$).
These results suggest that track accuracy alone does not
explain policy performance; the way motion representations
interact with visual features and action generation also
matters. The full model combines accurate future-track
prediction with effective action generation, although
these comparisons do not isolate a causal effect of
track accuracy on manipulation success.
\begin{table*}
    \centering
    \caption{\textbf{Per-task ablation results} on eight RoboTwin tasks. We report per-task average success rate (\%) across three evaluation seeds, with 50 rollouts per seed per task. Best results are shown in \textbf{bold} and second-best results are \underline{underlined}.}
    \label{tab:per_task_ablation}
    \scriptsize
    \setlength{\tabcolsep}{2.0pt}
    \begin{tabular*}{\textwidth}{@{}l@{\extracolsep{\fill}}ccccccccc@{}}
        \toprule
        & \textbf{\cellcolor{gray!15}Avg.$\uparrow$}
        & \textup{Handover}
        & \textup{Stack Blocks}
        & \textup{Stack Blocks}
        & \textup{Scan}
        & \textup{Place Bread}
        & \textup{Place Dual}
        & \textup{Pick Diverse}
        & \textup{Place Bread} \\
        \textbf{Variant}
        & \cellcolor{gray!15}(\%)
        & \textup{Block}
        & \textup{Two}
        & \textup{Three}
        & \textup{Object}
        & \textup{Skillet}
        & \textup{Shoes}
        & \textup{Bottles}
        & \textup{Basket} \\
        \midrule

        Action-only
        & \cellcolor{gray!15}61.4
        & 67.3 & 71.3 & 59.3 & 63.3
        & 48.0 & 43.3 & 74.0 & 64.7 \\

        Track regression
        & \cellcolor{gray!15}73.9
        & 94.7 & 83.3 & 80.0 & 59.3
        & 70.0 & 60.7 & 77.0 & 66.0 \\

        Track-conditioned action diffusion
        & \cellcolor{gray!15}\underline{76.3}
        & \underline{95.0} & 85.3 & 84.0 & 54.7
        & \textbf{75.0} & 59.0 & \textbf{87.0} & 70.0 \\

        w/o Visual--Track Concatenation
        & \cellcolor{gray!15}70.0
        & 81.3 & 78.0 & 70.0 & 60.0
        & 68.7 & \textbf{67.3} & 74.7 & 60.0 \\

        w/o 4D RoPE
        & \cellcolor{gray!15}74.1
        & 86.7 & \underline{90.7} & \underline{86.7}
        & \underline{66.0}
        & 67.3 & \underline{62.0} & 61.3 & \underline{72.0} \\

        \textbf{Full model}
        & \cellcolor{gray!15}\textbf{81.7}
        & \textbf{96.7} & \textbf{91.3} & \textbf{96.0}
        & \textbf{72.0}
        & \underline{72.7} & \underline{62.0}
        & \underline{85.3} & \textbf{77.3} \\

        \bottomrule
    \end{tabular*}
\end{table*}

\begin{table}[t]
\caption{\textbf{Future 3D track prediction} on the validation
sets of the eight RoboTwin tasks. ADE and FDE are computed
over moving patches and averaged across tasks.
Best results are shown in \textbf{bold}.}
\label{tab:track_prediction_appendix}
\centering
\small
\setlength{\tabcolsep}{3pt}
\begin{tabular}{@{}lcc@{}}
\toprule
Variant & ADE (mm) $\downarrow$ & FDE (mm) $\downarrow$ \\
\midrule
Track regression
& 11.1 & 14.8 \\
Track-conditioned action diffusion
& 9.9 & 12.5 \\
w/o Visual--Track Concatenation
& 11.9 & 15.3 \\
w/o 4D RoPE
& 13.2 & 17.2 \\
\textbf{Full model}
& \textbf{7.6} & \textbf{10.0} \\
\bottomrule
\end{tabular}
\end{table}

\subsection{Per-Task Easy-to-Hard Generalization Results}
Table~\ref{tab:per_task_easy_to_hard} reports per-task results
when transferring policies trained on Easy scenes to Hard scenes
with unseen textures and clutter, without fine-tuning.
Each policy is evaluated using three evaluation seeds with
50 rollouts per seed, and we report the mean and standard
deviation across these seeds.
\method achieves the highest success rate on seven of the eight
tasks, with an average of $17.9\%$ compared with $4.3\%$ for
the strongest baseline, GAP.
However, success rates remain low overall, and \method does
not succeed on Place Bread Skillet, highlighting the remaining
challenges of generalization under these visual changes.

\begin{table*}[t]
    \centering
    \caption{\textbf{Per-task easy-to-hard generalization} on eight RoboTwin tasks. All methods are trained on \emph{Easy} and evaluated on \emph{Hard} without fine-tuning.
    We report task success rate (\%) as mean $\pm$ standard deviation across three evaluation seeds. Best results are shown in \textbf{bold} and second-best results are \underline{underlined}.}
    \label{tab:per_task_easy_to_hard}
    \scriptsize
    \setlength{\tabcolsep}{2.0pt}
    \begin{tabular*}{\textwidth}{@{}l@{\extracolsep{\fill}}ccccccccc@{}}
        \toprule
        & \textbf{\cellcolor{gray!15}Avg.$\uparrow$}
        & Handover
        & Stack Blocks
        & Stack Blocks
        & Scan
        & Place Bread
        & Place Dual
        & Pick Diverse
        & Place Bread \\
        \textbf{Method}
        & \cellcolor{gray!15}(\%)
        & Block
        & Two
        & Three
        & Object
        & Skillet
        & Shoes
        & Bottles
        & Basket \\
        \midrule

        DP3~\cite{ze20243d}
        & \cellcolor{gray!15}1.7
        & \result{4.7}{2.5}
        & \result{\underline{1.3}}{0.9}
        & \result{0.0}{0.0}
        & \result{2.0}{1.6}
        & \result{0.0}{0.0}
        & \result{0.0}{0.0}
        & \result{2.7}{1.9}
        & \result{2.7}{2.5} \\

        DICP~\cite{xu2025imaginative}
        & \cellcolor{gray!15}3.9
        & \result{0.7}{0.9}
        & \result{0.7}{0.9}
        & \result{0.0}{0.0}
        & \result{\underline{4.7}}{2.5}
        & \bestresult{8.7}{3.4}
        & \result{3.3}{2.5}
        & \result{4.7}{3.4}
        & \result{\underline{8.0}}{6.5} \\

        GAP~\cite{xu2026gap}
        & \cellcolor{gray!15}4.3
        & \result{\underline{6.0}}{1.6}
        & \result{0.0}{0.0}
        & \result{0.0}{0.0}
        & \result{3.3}{2.5}
        & \result{\underline{2.0}}{2.8}
        & \result{\underline{8.0}}{1.6}
        & \result{\underline{9.3}}{1.9}
        & \result{6.0}{1.6} \\

        ATM~\cite{wen2023any}
        & \cellcolor{gray!15}0.0
        & \result{0.0}{0.0} & \result{0.0}{0.0} & \result{0.0}{0.0} & \result{0.0}{0.0} & \result{0.0}{0.0} & \result{0.0}{0.0} & \result{0.0}{0.0} & \result{0.0}{0.0} \\

        \textbf{\method}
        & \cellcolor{gray!15}\textbf{17.9}
        & \bestresult{12.0}{1.6}
        & \bestresult{10.7}{3.8}
        & \bestresult{22.0}{3.3}
        & \bestresult{13.3}{2.5}
        & \result{0.0}{0.0}
        & \bestresult{12.7}{3.4}
        & \bestresult{47.3}{6.6}
        & \bestresult{25.3}{5.2} \\

        \bottomrule
    \end{tabular*}
\end{table*}

\subsection{Additional Future Prediction Targets}
We further examine whether predicting richer future observations
provides additional benefits beyond 3D tracks.
Starting from the full \method model, we additionally predict
either future RGB observations or future 3D geometry features,
while retaining the same DiT backbone configuration and
evaluation protocol.

For future-image prediction, we encode RGB observations using
a frozen variational autoencoder (VAE) from Stable Diffusion XL
(SDXL) and supervise the predicted VAE latents.
For endpoint geometry prediction, we extract target features
using a frozen pretrained Pi3 encoder and apply supervision
directly in feature space. Although these features can be
decoded into a point map, the loss is computed on the features
rather than the decoded geometry.
In both variants, the additional latent or feature tokens are
incorporated into the DiT sequence, grounded by 4D RoPE, and
jointly denoised with actions and tracks.

Table~\ref{tab:future_prediction_ablation} compares these variants
on the same eight RoboTwin tasks used for ablation.
Adding future-image or geometry-feature prediction reduces
average success from $81.7\%$ to $79.4\%$ and $74.2\%$,
respectively. Under the evaluated configuration, these
additional prediction targets provide no further performance
benefit over joint action--track prediction alone.

\begin{table}[t]
\caption{\textbf{Effect of additional future prediction targets} on eight RoboTwin tasks. We report average success rates (\%) over three evaluation seeds, with 50 rollouts per seed per task. Best results are shown in \textbf{bold}.}
\label{tab:future_prediction_ablation}
\centering
\small
\setlength{\tabcolsep}{6pt}
\begin{tabular}{lccc}
\toprule
Future prediction target & \cellcolor{gray!15} Avg.$\uparrow$ & Sync. & Seq. \\
\midrule
3D tracks & \cellcolor{gray!15} \textbf{81.7} & \textbf{74.3} & \textbf{89.0} \\
3D tracks + future image & \cellcolor{gray!15} 79.4 & 71.8 & 87.0 \\
3D tracks + future 3D geometry & \cellcolor{gray!15} 74.2 & 64.5 & 83.9 \\
\bottomrule
\end{tabular}
\end{table}

\subsection{Alternative Future Representations}
We compare different future prediction targets within our
framework on the same eight RoboTwin tasks.
The full-horizon 2D track variant replaces 3D displacement
targets with 2D pixel displacements over the same prediction
horizon, with the prediction head adjusted accordingly.
We evaluate this variant both with and without 4D RoPE.
When enabled, 4D RoPE retains the same depth-derived world-space
anchors as the full model, even though the prediction targets
are in image space.

The endpoint 3D displacement variant predicts only the
displacement at the final horizon step, whereas the full model
predicts displacements at all future steps.
Both variants retain patch-aligned visual--track fusion.
For the endpoint geometry variant, we replace the track
prediction target with Pi3 features extracted by a frozen
pretrained encoder. Patch-level noisy geometry features are
concatenated with their corresponding visual features,
and supervision is applied directly in feature space
rather than to the decoded point map.
All variants jointly denoise their respective future
representations with actions.

As shown in Table~\ref{tab:motion_representation_ablation},
adding 4D RoPE to full-horizon 2D tracks improves average
success from $76.1\%$ to $79.5\%$, supporting the value of
spatiotemporal grounding across modalities even when motion
is predicted in image space.
Endpoint 3D displacements and full-horizon 3D tracks achieve
comparable success rates of $81.2\%$ and $81.7\%$, respectively,
while endpoint 3D geometry achieves $82.8\%$.
Performance across these three 3D prediction targets falls
within $1.6$ percentage points, demonstrating that our joint
prediction architecture accommodates different future
representations with comparable manipulation success.
Our track-based formulation achieves comparable performance
using depth-derived spatial anchors, without the additional
Pi3 feature extraction used by the geometry variant.

We further compare these representations under an Easy-to-Hard
visual shift, where all variants are trained on Easy scenes
and evaluated on Hard scenes without fine-tuning.
Despite their similar in-distribution performance,
full-horizon 3D tracks achieve higher Easy-to-Hard success
($17.9\%$) than endpoint displacements ($12.2\%$) and endpoint
geometry ($10.3\%$), as shown in
Table~\ref{tab:per_task_representation_generalization}.
These results suggest that full-horizon 3D tracks support
better generalization under the evaluated visual shift.
One possible explanation is that endpoint geometry features
encode surrounding scene structure and appearance, potentially
making them more sensitive to changes in background and clutter.
Endpoint displacements focus on point motion but supervise
only the net change at the final step.
Full-horizon tracks additionally supervise intermediate motion,
which may encourage the model to learn task-relevant motion
patterns that transfer across visual conditions.
These explanations remain hypotheses rather than mechanisms
isolated by the current experiments.

\begin{table}[t]
\caption{\textbf{Comparison of alternative future representations} on eight RoboTwin tasks. We report average success rates (\%) over three evaluation seeds, with 50 rollouts per seed per task. Best results are shown in \textbf{bold} and second-best results are \underline{underlined}.}
\label{tab:motion_representation_ablation}
\centering
\small
\setlength{\tabcolsep}{5pt}
\begin{tabular}{lccc}
\toprule
Future prediction target & \cellcolor{gray!15} Avg.$\uparrow$ & Sync. & Seq. \\
\midrule
2D tracks (full horizon)
& \cellcolor{gray!15} 76.1 & 67.5 & 84.7 \\

2D tracks + 4D RoPE (full horizon)
& \cellcolor{gray!15} 79.5 & 70.2 & 88.8 \\

3D geometry (endpoint)
& \cellcolor{gray!15} \textbf{82.8} & \textbf{76.7} & 88.9 \\

3D point displacements (endpoint)
& \cellcolor{gray!15} 81.2 & 73.0 & \textbf{89.4} \\

\textbf{3D tracks (full horizon, ours)}
& \cellcolor{gray!15} \underline{81.7} & \underline{74.3} & \underline{89.0} \\
\bottomrule
\end{tabular}
\end{table}

\begin{table*}[t]
    \centering
    \caption{\textbf{Per-task easy-to-hard generalization with alternative future representations} on eight RoboTwin tasks. We report task success rate (\%) as mean $\pm$ standard deviation across three evaluation seeds. Best results are shown in \textbf{bold}.}
    \label{tab:per_task_representation_generalization}
    \scriptsize
    \setlength{\tabcolsep}{2.0pt}
    \begin{tabular*}{\textwidth}{@{}l@{\extracolsep{\fill}}ccccccccc@{}}
        \toprule
        & \textbf{\cellcolor{gray!15}Avg.$\uparrow$}
        & Handover
        & Stack Blocks
        & Stack Blocks
        & Scan
        & Place Bread
        & Place Dual
        & Pick Diverse
        & Place Bread \\
        \textbf{Future prediction target}
        & \cellcolor{gray!15}(\%)
        & Block
        & Two
        & Three
        & Object
        & Skillet
        & Shoes
        & Bottles
        & Basket \\
        \midrule

        2D tracks (full horizon)
        & \cellcolor{gray!15}9.3
        & \result{2.7}{2.5}
        & \result{16.7}{3.4}
        & \bestresult{22.0}{4.3}
        & \result{3.3}{0.9}
        & \result{0.0}{0.0}
        & \result{5.3}{2.5}
        & \result{16.0}{1.6}
        & \result{8.7}{2.5} \\

        2D tracks + 4D RoPE (full horizon)
        & \cellcolor{gray!15}8.8
        & \result{0.0}{0.0}
        & \result{12.7}{3.4}
        & \result{3.3}{3.4}
        & \result{8.7}{5.2}
        & \result{0.0}{0.0}
        & \result{6.7}{1.9}
        & \result{31.3}{8.2}
        & \result{7.3}{3.4} \\

        3D geometry (endpoint)
        & \cellcolor{gray!15}10.3
        & \result{2.7}{0.9}
        & \result{11.3}{5.0}
        & \result{1.3}{1.9}
        & \result{10.7}{2.5}
        & \result{0.0}{0.0}
        & \result{10.7}{4.7}
        & \result{28.7}{5.2}
        & \result{17.3}{5.2} \\

        3D point displacements (endpoint)
        & \cellcolor{gray!15}12.2
        & \result{7.3}{1.9}
        & \bestresult{26.7}{8.2}
        & \result{5.3}{1.9}
        & \bestresult{13.3}{3.8}
        & \result{0.0}{0.0}
        & \result{2.7}{0.9}
        & \result{31.3}{2.5}
        & \result{11.3}{2.5} \\

        \textbf{3D tracks (full horizon, ours)}
        & \cellcolor{gray!15}\textbf{17.9}
        & \bestresult{12.0}{1.6}
        & \result{10.7}{3.8}
        & \bestresult{22.0}{3.3}
        & \bestresult{13.3}{2.5}
        & \result{0.0}{0.0}
        & \bestresult{12.7}{3.4}
        & \bestresult{47.3}{6.6}
        & \bestresult{25.3}{5.2} \\

        \bottomrule
    \end{tabular*}
\end{table*}

\subsection{Additional Implementation Details}
\label{app:implementation}

\subsubsection{Architecture and Optimization}
For real-world experiments, the policy uses a single
head-camera RGB view with an observation horizon of $H_o=1$.
We crop regions outside the tabletop workspace, resulting in
input images of $168\times322$ pixels (height $\times$ width).
DINOv2~\cite{oquab2023dinov2} encodes these images with a patch
size of $14$, yielding a $12\times23$ patch grid and $N=276$
track queries, one per visual patch.
Both the action and track prediction horizons are set to
$H_a=H_p=16$.

\method uses a DiT with $4$ Transformer blocks, a hidden
dimension of $1024$, and $8$ attention heads.
We optimize the model with AdamW using a learning rate of
$10^{-4}$, $(\beta_1,\beta_2)=(0.9,0.999)$, weight decay
$10^{-6}$, and a batch size of $64$.
The learning-rate schedule consists of $500$ warm-up steps
followed by cosine decay.

\subsubsection{Diffusion and Loss Hyperparameters}
We use clean-sample prediction (\texttt{prediction\_type=sample})
for both action and track diffusion, supervising the outputs
$\widehat{Y}_A$ and $\widehat{Y}_P$ with normalized clean targets:
\begin{equation}
Y_A = \widetilde{A}_t^0,
\qquad
Y_P = \widetilde{P}_t^0.
\end{equation}
Both branches use the same cosine noise schedule
(\texttt{squaredcos\_cap\_v2}) with $K=100$ training diffusion
timesteps and $10$ DDIM denoising steps at inference.

We set $\lambda_{\mathrm{track}}=10^{-4}$,
$w_{\min}=0$, and $\tau=0.01\,\mathrm{m}$, with sigmoid
steepness $\kappa=5/\tau=500\,\mathrm{m}^{-1}$.
The track loss sums squared errors over the prediction
horizon and three spatial coordinates, weighted by
$w_{b,i}$ for query $i$ in batch element $b$, and divides by
$3\max(\sum_{b,i}w_{b,i},10^{-6})$.
It is therefore not averaged over the prediction horizon.

\subsubsection{Track Validity and Invalid-Query Supervision}
For simulator-generated track supervision, a patch query is
valid if its center has a positive, finite depth value sampled
by bilinear interpolation, and its segmentation ID belongs
to a tracked rigid body. Tracked bodies include objects and
robot arm links. Queries with invalid depth or segmentation
IDs outside this set are assigned zero displacement targets.

For each valid query, we transform its anchor-frame 3D position
into the owning body's local frame and compute future positions
using that body's ground-truth poses. Validity is determined
at the anchor frame and propagated across the prediction
horizon; future visibility and occlusion are not re-evaluated.
For prediction steps beyond the episode boundary, we reuse
the final recorded pose.

We normalize 3D track displacements as
$\widetilde{P}_t^0=P_t^0/\sigma_{\mathrm{rms}}$, where
$\sigma_{\mathrm{rms}}$ is the root mean square of all valid
displacement components across samples and prediction steps
in each training dataset. This scalar scale is shared across
the three spatial dimensions, with no mean subtraction,
so zero displacement targets remain zero after normalization.
Invalid queries are excluded from scale estimation but included
in the movement-weighted track loss without validity masking.
Under the specified loss hyperparameters, they receive a weight
of $\operatorname{sigmoid}(-5)\approx0.0067$ and therefore
contribute weak zero-displacement supervision.

\subsubsection{Training Duration and Checkpoint Selection}
DP and DP3 are trained for $600$, $3000$ epochs, respectively. \method, DICP, and GAP are each trained for $200$ epochs. For ATM, the Track Transformer is trained for $100$ epochs, while the ATM Diffusion Policy is trained for $200$ epochs. For every method, we use the checkpoint from the final training epoch for evaluation.
\subsubsection{4D-RoPE Implementation}
\section{Spatiotemporal Grounding with 4D RoPE}
\label{app:4d_rope}

We apply four-dimensional rotary positional encoding (4D RoPE) to incorporate relative spatial and temporal relationships into multimodal attention. Each grounded token is associated with a coordinate $\mathbf{c} = (x,y,z,u)$, where $(x,y,z)$ denotes its position in a common world frame, measured in meters, and $u$ denotes its physical timestep index rather than the diffusion timestep.

\paragraph{Axis allocation and coordinate scaling}
Each attention head has dimension $128$, which we divide into four groups of $32$ dimensions for the $x$, $y$, $z$, and temporal axes. RoPE is applied independently to each group of the query and key vectors. For an axis coordinate $c$, the rotation angle for the $j$-th pair of feature dimensions is
\begin{equation}
    \phi_j(c)
    =
    \frac{c}{s}\,b^{-2j/d},
    \qquad j=0,\ldots,d/2-1,
\end{equation}
where $d=32$ is the dimension allocated to each axis. We use a frequency base of $b=100$ for all four axes, a spatial scale of $s_{\mathrm{xyz}}=0.01\,\mathrm{m}$, and a temporal scale of $s_u=16$. These rotations encode relative coordinate offsets in query--key dot products.

\paragraph{Token coordinates}
Visual patches and their associated track queries are anchored at the corresponding current 3D patch centers and assigned temporal index $u=0$. The fused visual--track tokens inherit these coordinates. Each arm-state token is anchored at that arm's current end-effector position, also with $u=0$. 
Action tokens are assigned temporal indices $u=h$ for
$h=1,\ldots,H_a$ and spatial coordinates computed from the
current noisy action sequence, as described below. Special tokens, including the CLS token when present and the diffusion-timestep token, are excluded from rotary encoding.

Each track token represents the complete future trajectory of one query point rather than an individual future timestep. Its temporal coordinate therefore identifies the current reference frame from which the trajectory is predicted. The ordered future displacements are encoded within the track embedding, while temporally indexed action tokens provide timestep-specific interactions through attention.

\paragraph{Noise-dependent action grounding}
To obtain spatial anchors for action tokens, we combine the current end-effector position with a position derived from the noisy action at the current diffusion step. Let $\mathbf{g}_a$ denote the current world-frame end-effector position of arm $a$, and let $\widetilde{\mathbf{a}}^{\,k}_{h,a}$ denote its normalized noisy action at future step $h$ and diffusion step $k$. We define
\begin{equation}
    \mathbf{r}^{\,k}_{h,a}
    =
    \Phi_a\!\left(
        \widetilde{\mathbf{a}}^{\,k}_{h,a}
    \right),
\end{equation}
where $\Phi_a$ applies the inverse action normalization and
extracts the absolute end-effector position of arm $a$.
Both arms' positions are expressed in the same world frame. The spatial anchor is
\begin{equation}
    \mathbf{x}^{\,k}_{h,a}
    =
    (1-\rho_k)\mathbf{g}_a
    +
    \rho_k\mathbf{r}^{\,k}_{h,a},
    \qquad
    \rho_k = 1-\frac{k}{K},
\end{equation}
where $k\in\{0,\ldots,K-1\}$ denotes the scheduler timestep
on the training diffusion grid, rather than the DDIM
iteration index. At inference, we use the scheduler-selected
timesteps to compute $\rho_k$.

At high noise levels, the anchor remains close to the current end-effector position, reducing the influence of noisy action coordinates on rotary encoding. As denoising progresses, the anchor increasingly follows the position derived from the evolving action sequence. We recompute these anchors at every denoising step and use the same construction during training and inference. The anchors are computed from the current noisy actions, without using clean ground-truth future actions.
\section{Additional ablations}

\end{document}